# Are you tough enough? Framework for Robustness Validation of Machine Comprehension Systems

**Barbara Rychalska**[*]
Warsaw University of Technology
b.rychalska@mini.pw.edu.pl

**Dominika Basaj**[*]
Warsaw University of Technology
Tooploox
d.basaj@mini.pw.edu.pl

**Przemyslaw Biecek**
Warsaw University of Technology
p.biecek@mini.pw.edu.pl

## Abstract

Deep Learning NLP domain lacks procedures for the analysis of model robustness. In this paper we propose a framework which validates robustness of any Question Answering model through model explainers. We propose that a robust model should transgress the initial notion of semantic similarity induced by word embeddings to learn a more human-like understanding of meaning. We test this property by manipulating questions in two ways: swapping important question word for 1) its semantically correct synonym and 2) for word vector that is close in embedding space. We estimate importance of words in asked questions with Locally Interpretable Model Agnostic Explanations method (LIME). With these two steps we compare state-of-the-art Q&A models. We show that although accuracy of state-of-the-art models is high, they are very fragile to changes in the input. Moreover, we propose 2 adversarial training scenarios which raise model sensitivity to true synonyms by up to 7% accuracy measure. Our findings help to understand which models are more stable and how they can be improved. In addition, we have created and published a new dataset that may be used for validation of robustness of a Q&A model[1].

## 1 Introduction

Up-to-date advancements in natural language processing show that it is possible to achieve high accuracy on tasks that from human point of view require language understanding. However, recent research on adversarial examples revealed shortcomings of neural network models, despite their great performance on selected datasets [11]. Most adversarial examples studies in Question Answering (Q&A) task focus on designing sophisticated attacks that prove overstability understood as an inability of the model to distinguish sentences that answer the question from sentences that have words in common with the question [4]. [6] and [3] also show that the whole question itself is not crucial to get the right answer. Despite growing research in the area of adversarial attacks, little is still known about the source of lacking robustness of Q&A models and effective ways of overcoming this problem. In order to come up with a solution it is crucial to understand the inner workings of task specific architectures and benchmark their performance on adversarial attacks.

---

[*]Equal contribution
[1]*https://github.com/MI2DataLab/nlp_interpretability_framework*

In this work we propose a model agnostic framework that checks stability of three state-of-the-art Q&A architectures in terms of their ability to respond to semantically similar questions. We formulate two aspects which are subject to tests: semantic input stability (relying on true semantics) and numerical input stability (purely induced by similarity between word embeddings). These terms are explained later on in the article. We state that a robust and reliable Q&A model should be invariant to changes in the input until they induce semantic changes. Having said that, we state that a stable model must display higher semantic input stability measure than numerical input stability measure. We claim that a robust Q&A model should possess high sensitivity to true semantics, as opposed to closeness between word embeddings. Otherwise, attacks based on commonly accessible sets of embeddings become possible, including antonymous questions, as antonyms are often close in embedding space.

We make the following contributions:

- We investigate robustness of the model in the face of semantic and numerical changes to the input.
- We use output of Locally Interpretable Model Agnostic Explanations [10] to create attacks.
- We offer 2 approaches to adversarial training which increase model sensitivity to true semantic differences by a maximum of 7% in accuracy.
- We release a collection of 1500 semantically coherent questions from SQuAD dataset [9] preprocessed by LIME and human annotators, that we used in this study, as a reference dataset for further works on this problem.

In section 2 we explain measures and tools that we used in the study. In section 3 we conduct experiments on three popular Q&A architectures, then we introduce two modifications to one of the models in order to increase ability to answer semantically similar questions and finally we refer to related work and conclude.

## 2 Robustness measures

**Importance of question words.** Previous studies have shown sensitivity of the model to changes in the input. Based on this notion, we use the popular LIME (Locally Interpretable Model Agnostic Explanations) framework to assess the contribution of each question word and construct adversaries later on, by replacing the most important word. LIME fits a linear model in local surrounding of each example and estimates coefficients that serve as a importance measure in the process of model interpretability. We call the most important word the *keyword*.

**Numerical input stability.** Deep learning models usually have input words represented in the form of embedded vectors, which exhibit proximity to other words, which are not necessarily synonyms. This results from their creation procedure, which represents a word by its neighbors. In GloVe [7] and Fasttext [1] antonyms can be found to be close neighbors, e.g. **victory** is located near to **defeat**, as well as **up** to **down** (0.85 cosine similarity in GloVe, they are nearest neighbors). Likewise, other kinds of similarity which bring words close together in embedding space can be identified, such as common hypernym, e.g. **cat** is very close to **dog** and **puppy**. Overstability to numerical inputs should be thus regarded as major vulnerability to adversarial attacks.

We use the idea of similarity in embedding space to find words that are the closest to the *keyword* in SQuAD questions. We swap *keywords* with closest words and make predictions. GloVe embeddings are used for extraction of closest words, since they were the input used to train the tested models. Overall, we find that 37% of our created questions can be seen as potentially vulnerability-inducing, containing antonyms or other severe meaning modifications with regard to original questions.

**Semantic input stability.** As stated above, robust model should be invariant to semantic changes. Thus, we disregard common word embeddings and construct genuinely synonymous sentences based on Wordnet [5] synonyms end context dependent ELMO [8] embeddings that are later on validated by human annotators and corrected. This manually improved dataset is released for the community. The procedure together with comparison with numerical input stability measure allows us to inspect whether neural network **understands** the whole question or simply relies on numerical representation of *keywords*.

Table 1: Example alterations for question "How many people lived in Warsaw in 1939?"

| Type of alteration | Example sentence |
|---|---|
| Original | How many **people** lived in Warsaw in 1939? |
| Synonym | How many **citizens** lived in Warsaw in 1939? |
| Numeric | How many **those** lived in Warsaw in 1939? |
| Random | How many**random** lived in Warsaw in 1939? |

Table 2: Performance and robustness metrics for selected architectures

| Architecture | Numeric accuracy | Synonym accuracy | Rand accuracy | Original EM | Original F1 |
|---|---|---|---|---|---|
| DrQA | 0.69 | 0.64 | 0.43 | 0.69 | 0.78 |
| BiDAF | 0.70 | 0.65 | 0.48 | 0.68 | 0.77 |
| QANet | 0.72 | 0.70 | 0.51 | 0.69 | 0.79 |

# 3 Experiments

We investigate three popular architectures: DrQA [2], BiDAF [12] and QANet [14]. Each model is tested on a prepared dataset, which comprises of examples of contexts, questions and answers extracted from SQuAD dataset. Each question is modified so that the most important word computed with LIME is swapped for 1. its synonym appropriate in the given context (**syn-Dataset**) 2. a word which has the shortest cosine distance in GloVe embedding space (**num-Dataset**) 3. a word *random* (**random-Dataset**). Table 1 shows an example of the resulting questions. In total, we produce 1500 test examples, with 500 generated by each of the 3 algorithms. Note that modifications induced by various algorithms vary since they attend to different words in the question (as measured by LIME), so in fact we release 3 separate datasets. In our experiments we only test on examples which scored a correct answer in each dataset, treating wrongly tagged examples as of ambiguous merit for this task. A merge of these three datasets can be seen as an approximation of our original datasets to be used with any new model other than the ones we test here.

As shown in table 2, stability with regard to both semantic and numeric changes varies between models. In general however, we find that all models respond better to changes induced by num-Dataset than syn-Dataset, which is an unwelcome result. Indeed, all tested models are found to be overstable to numeric manipulations while recognizing true synonyms at a lower rate. Moreover, there is a big difference in performance once we swap *keyword* with random words which are on average more distant to the *keyword* in embedding space. This heavy decrease in performance also proves that indeed, models are overstable once attacked with words close to the *keyword*.

# 4 Adversarial training

In the face of our findings, a much-needed next step is to render models more robust by increasing their ability to give right answer to the semantically similar questions. In order to increase this aspect of model robustness we introduce two separate changes to the algorithm and choose to retrain DrQA model, which turned out to be weak in terms of semantic similarity recognition. New models are validated on all datasets. These changes are of model-agnostic nature and can be implemented in any algorithm.

Table 3: Results of our adversarial training scenarios: REM (removal of random words) with 2 and 3 modified training examples added for each original example, and GRAD (propagation of gradients to inputs) on 10,000 and 30,000 top words.

| Measure | benchmark | REMx2 | REMx3 | GRADx10k | GRADx30k |
|---|---|---|---|---|---|
| Numeric accuracy | 0.69 | 0.77 (+0.08) | 0.77 (+0.08) | 0.76 (+0.07) | 0.76 (+0.07) |
| Synonym accuracy | 0.64 | **0.71 (+0.07)** | 0.70 (+0.06) | 0.70 (+0.06) | 0.69 (+0.05) |

**Removing words from training dataset (REM).** Inspired by logic behind dropout [13] we train Q&A models with enriched training dataset. Each question is present in multiple forms - one question with all words preserved and 2 or 3 questions with one random word removed. This enrichment of the dataset enables to overcome phenomenon that may be the reason behind the success of adversarial attacks - heavy dependence on individual words in the question. Thanks to this procedure we increase the performance on syn-Dataset in comparison to training only on original questions. However, it must be noted that although this improvement is a step towards answering semantically similar questions, it does not overcome the problem of higher numeric input stability than semantic input stability, defined in section 2.

**Propagating gradients back to input embeddings (GRAD).** Intuitively, a part of the Q&A model's knowledge of synonymy acquired during training can be transmitted to embeddings themselves. Fine-tuning is sometimes done to increase model performance, and may be abandoned when no rise in performance is observed. We conclude that fine-tuning the embeddings can increase the semantic stability, independently from influencing model performance. Similarly as in REM procedure, we obtain a rise in semantic synonym recognition, coupled with a rise in numeric accuracy.

To check whether the increased numeric accuracy scores were indeed a problem in our adversarial settings, we tested which percentage of semantically incorrect num-Dataset examples receive a negative score from our models. For base model, 37% problematic examples received decision change with respect to correct original examples. With scores 38%, 36% for GRADx10 and GRADx30 respectively, these models suffer less overstability to numeric modifications with regard to base model, while REM models (34%, 30% for REMx2 and REMx3 respectively) did exhibit increased acceptance to manipulations in numeric representations.

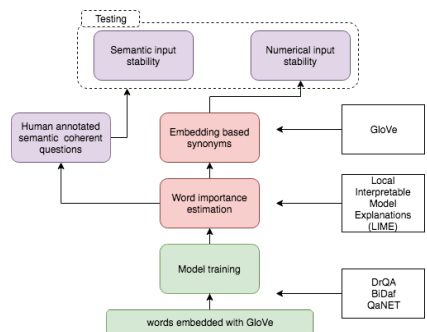

Figure 1: Schema of robustness validation framework.

Both methods influenced model robustness without a noticeable difference in model quality measured in traditional metrics - EM and F1. It suggests that models do need specific, established procedures for testing robustness, which should be conducted independently from traditional performance tests, since these scores are not necessarily related.

## 5 Related work

Researching true semantics with respect to numeric manipulations is a novel topic, although recent research is rich with papers dealing with manipulating semantics for Q&A systems. [11] show effectiveness of adversarial attacks on Q&A system which drop performance without significant semantic change. [4] show a keyword detection method in Q&A using Integrated Gradients, although they do not venture on to create attacks using important words in Q&A task. [15] generate paraphrases of whole adversarial questions and test model robustness on these grammatically correct and semantically similar adversaries. In our work, apart from studying semantic similarity on a more detailed level (individual words), we propose an augmentation framework. [14] attempt to generate adversarial examples in a non-gradient based way, in order to be able able to attack a black-box model, a problem which we bypass using model-agnostic explainer.

## 6 Conclusion

In our study we focused on one particular aspect of model robustness - ability to answer semantically coherent questions. We show that performance of all tested models decreases once we change important question words indicated by LIME. However, we observe that models have higher performance once we swap tokens in questions for words close in their embedding space defined by word vectors they were trained on, in comparison to accuracy obtained by asking semantically correct questions. We manage to increase ability of the model to answer this kind of questions but it does not mean that they understand semantics - it is reflected in increased accuracy of questions with GloVe embeddings once we test our newly trained models on num-Dataset.

We show that the reason behind the success of some adversarial examples lies in the way input words are represented. Popular embeddings do not include knowledge about *real* meaning of the words, but rather incorporate knowledge about their context. Our work serves as a starting point for future research on more semantically-conscious representation of words.

## Acknowledgements

This research was supported by the Polish National Science Centre DALEX grant no. 2017/27/B/ST6/01307.

## References

[1] Piotr Bojanowski, Edouard Grave, Armand Joulin, and Tomas Mikolov. Enriching word vectors with subword information. *Transactions of the Association for Computational Linguistics*, 5:135–146, 2017.

[2] Danqi Chen, Adam Fisch, Jason Weston, and Antoine Bordes. Reading Wikipedia to answer open-domain questions. In *Association for Computational Linguistics (ACL)*, 2017.

[3] Shi Feng, Eric Wallace, Mohit Iyyer, Pedro Rodriguez, Alvin Grissom II, and Jordan L. Boyd-Graber. Right answer for the wrong reason: Discovery and mitigation. *CoRR*, abs/1804.07781, 2018.

[4] Robin Jia and Percy Liang. Adversarial examples for evaluating reading comprehension systems. In *Proceedings of the 2017 Conference on Empirical Methods in Natural Language Processing*, pages 2021–2031. Association for Computational Linguistics, 2017.

[5] George A. Miller. Wordnet: A lexical database for english. *Commun. ACM*, 38(11):39–41, November 1995.

[6] Pramod Kaushik Mudrakarta, Ankur Taly, Mukund Sundararajan, and Kedar Dhamdhere. Did the model understand the question? *CoRR*, abs/1805.05492, 2018.

[7] Jeffrey Pennington, Richard Socher, and Christopher D. Manning. Glove: Global vectors for word representation. In *Empirical Methods in Natural Language Processing (EMNLP)*, pages 1532–1543, 2014.

[8] Matthew E. Peters, Mark Neumann, Mohit Iyyer, Matt Gardner, Christopher Clark, Kenton Lee, and Luke Zettlemoyer. Deep contextualized word representations. In *Proc. of NAACL*, 2018.

[9] Pranav Rajpurkar, Jian Zhang, Konstantin Lopyrev, and Percy Liang. Squad: 100, 000+ questions for machine comprehension of text. In *EMNLP*, 2016.

[10] Marco Túlio Ribeiro, Sameer Singh, and Carlos Guestrin. "why should I trust you?": Explaining the predictions of any classifier. *CoRR*, abs/1602.04938, 2016.

[11] Marco Tulio Ribeiro, Sameer Singh, and Carlos Guestrin. Semantically equivalent adversarial rules for debugging nlp models. In *Proceedings of the 56th Annual Meeting of the Association for Computational Linguistics (Volume 1: Long Papers)*, pages 856–865. Association for Computational Linguistics, 2018.

[12] Min Joon Seo, Aniruddha Kembhavi, Ali Farhadi, and Hannaneh Hajishirzi. Bidirectional attention flow for machine comprehension. *CoRR*, abs/1611.01603, 2016.

[13] Nitish Srivastava, Geoffrey Hinton, Alex Krizhevsky, Ilya Sutskever, and Ruslan Salakhutdinov. Dropout: A simple way to prevent neural networks from overfitting. *Journal of Machine Learning Research*, 15:1929–1958, 2014.

[14] Adams Wei Yu, David Dohan, Minh-Thang Luong, Rui Zhao, Kai Chen, Mohammad Norouzi, and Quoc V. Le. Qanet: Combining local convolution with global self-attention for reading comprehension. *CoRR*, abs/1804.09541, 2018.

[15] Zhengli Zhao, Dheeru Dua, and Sameer Singh. Generating natural adversarial examples. In *International Conference on Learning Representations*, 2018.6